\documentclass{article}

\usepackage{microtype}
\usepackage{graphicx}
\usepackage{subfigure}
\usepackage{booktabs} 
\usepackage{microtype}
\usepackage{graphicx}
\usepackage{subfigure}
\usepackage{booktabs} 
\usepackage{comment}
\usepackage{amsfonts}
\usepackage[hyphens]{url}  
\usepackage{graphicx} 
\usepackage{amsmath,amsthm,amsfonts,amssymb}
\usepackage{enumerate}
\usepackage{mathabx}
\usepackage{multirow}
\usepackage{epstopdf}
\usepackage{enumitem}
\usepackage{bbm}
\usepackage{bm}
\usepackage{rotating}
\usepackage{hyperref}

\usepackage[accepted]{icml2023}

\usepackage{amsmath}
\usepackage{amssymb}
\usepackage{mathtools}
\usepackage{amsthm}

\usepackage[capitalize,noabbrev]{cleveref}

\theoremstyle{plain}
\newtheorem{theorem}{Theorem}[section]
\newtheorem{proposition}[theorem]{Proposition}

\theoremstyle{definition}

\newtheorem{assumption}[theorem]{Assumption}
\theoremstyle{remark}

\usepackage[textsize=tiny]{todonotes}

\begin{document}

\twocolumn[
\icmltitle{A Universal Unbiased Method for Classification from Aggregate Observations}



\icmlsetsymbol{equal}{*}

\begin{icmlauthorlist}
\icmlauthor{Zixi Wei}{cqu}
\icmlauthor{Lei Feng}{ntu,riken}
\icmlauthor{Bo Han}{hkbu,riken}
\icmlauthor{Tongliang Liu}{mbzuai,usyd,riken}
\icmlauthor{Gang Niu}{riken}
\icmlauthor{Xiaofeng Zhu}{uestc}
\icmlauthor{Heng Tao Shen}{uestc}
\end{icmlauthorlist}
\icmlaffiliation{cqu}{College of Computer Science, Chongqing University, China}
\icmlaffiliation{ntu}{School of Computer Science and Engineering, Nanyang Technological University, Singapore}
\icmlaffiliation{riken}{RIKEN Center for Advanced Intelligence Project, Japan}
\icmlaffiliation{hkbu}{Department of Computer Science, Hong Kong Baptist University, China}
\icmlaffiliation{mbzuai}{Mohamed bin Zayed University of Artificial Intelligence, United Arab Emirates}
\icmlaffiliation{usyd}{Sydney AI Centre, School of Computer Science, The University of Sydney, Australia}
\icmlaffiliation{uestc}{School of Computer Science and Engineering, University of Electronic Science and Technology of China, China}

\icmlcorrespondingauthor{Lei Feng}{lfengqaq@gmail.com}

\icmlkeywords{Machine Learning, ICML}

\vskip 0.3in
]



\printAffiliationsAndNotice{}  

\begin{abstract}
In conventional supervised classification, true labels are required for \emph{individual} instances. However, it could be prohibitive to collect the true labels for individual instances, due to privacy concerns or unaffordable annotation costs. This motivates the study on \emph{classification from aggregate observations} (CFAO), where the supervision is provided to \emph{groups} of instances, instead of individual instances. CFAO is a generalized learning framework that contains various learning problems, such as multiple-instance learning and learning from label proportions. The goal of this paper is to present a novel \emph{universal} method of CFAO, which holds an \emph{unbiased estimator} of the classification risk for \emph{arbitrary losses}---previous research failed to achieve this goal. Practically, our method works by weighing the importance of each label for each instance in the group, which provides purified supervision for the classifier to learn. Theoretically, our proposed method not only guarantees the \emph{risk consistency} due to the unbiased risk estimator but also can be compatible with arbitrary losses. Extensive experiments on various problems of CFAO demonstrate the superiority of our proposed method.
\end{abstract}

\section{Introduction}
\emph{Classification} is one of the most frequently encountered problems in machine learning \cite{kotsiantis2006machine}. Over the past ten years, deep learning models have achieved promising performance on various classification tasks, while such a success heavily relies on a large number of high-quality labels for \emph{individual} instances \cite{jordan2015machine}. In many real-world scenarios, it could be prohibitive to collect strong supervision information for individual instances, due to privacy issues, confidentiality concerns, or unaffordable annotation costs. These challenges of individual annotations motivate us to consider the supervision for \emph{groups} of instances, instead of individual instances. 

The supervision of groups of instances can be feasible in many realistic applications. For example, to protect the information of individual data points, some summary statistics of groups of data could be disclosed. For the drug activity prediction task \cite{dietterich1997solving}, individual annotations are inaccessible, while group annotations are provided. Besides, collecting the supervision information of groups of instances could incur much fewer costs when the annotations costs are unaffordable for individual instances \cite{zhou2018brief}. In addition, making predictions about individual-level behavior based on group-level data also has gained much attention from other fields, such as ecological inference \cite{schuessler1999ecological,flaxman2015supported} and preference learning \cite{furnkranz2003pairwise,chu2005preference}.

The feasibility of group supervision motivates us to investigate the task of \emph{classification from aggregate observations} (CFAO), where we aim to learn a classifier with only supervision on groups of instances. In CFAO, training data are represented by groups of instances and we can only observe the aggregate information of the group. CFAO can be considered as a general learning framework, which contains various learning problems with different types of aggregate information. A well-known problem belonging to CFAO is \emph{multiple-instance learning} \cite{maron1997framework,carbonneau2018multiple}, where the aggregate information is whether the group has at least one positive instance. Another typical problem is \emph{learning from label proportions} \cite{yu2013proptosvm,scott2020learning}, where the aggregate information is the proportion of instances from each class in the group. In recent years, CFAO has received increasing attention and some interesting problems of CFAO have been investigated. \citet{bao2017classification} studied \emph{classification from pairwise similarities}, where the aggregate information is whether two instances in the group belong to the same class (similar) or not (dissimilar). \citet{cui2020classification} studied \emph{classification from triplet comparisons}, where the aggregate information is whether one instance is more similar to the other one, compared with the third one.
\begin{table*}[!t]
\centering
\caption{A brief introduction for typical examples of classification from aggregate observations. For a group of examples $(\bm{x}_{1:m},y_{1:m})$, the aggregate label is generated by $z=g(y_{1:m})$.}
\label{Examples}
\resizebox{1.0\textwidth}{!}{
\setlength{\tabcolsep}{2.5mm}{
\begin{tabular}{llll}
\toprule
Classification from $\cdots$     & Size $m$          & Aggregate Information& Aggregate Function $g$                                    \\ \midrule
pairwise similarity & $m=2$        & if $y_1$ and $y_2$ belong to same class or not&$g(y_1,y_2)=\mathbb{I}[y_1=y_2]$            \\ \midrule
triplet comparison &
  $m=3$ &
  if $d(y_1,y_2)$ is smaller than $d(y_1,y_3)$ & $g(y_{1:3})=\mathbb{I}[d(y_1,y_2)<d(y_1,y3)]$ \\ \midrule
multiple instances         & $m \geq 2$ & if at least one positive label exits in $y_{1:m}$ ($k=2$) &
$g(y_{1:m})=\max(y_{1:m})$\\ \midrule
label proportion        & $m \geq 2$ & proportion of data from each class in the group & $g_j(y_{1:m})=({\sum_{i=1}^m\mathbb{I}[y_i=j]})/{m}$          \\ \midrule
ordinal rank         & $m=2$        & if $y_1$ is larger than $y_2$, i.e., $y_1 \geq y_2$. & $g(y_1,y_2)=\mathbb{I}[y_1>y_2]$         \\ \bottomrule
\end{tabular}
}
}
\end{table*}

The goal of this paper is to propose a \emph{universal} approach that can be applied to various types of aggregate observations for CFAO. To the best of our knowledge, there is currently only one universal approach \cite{zhang2020learning} that can be used in various problems of CFAO. This approach is based on maximum likelihood estimation and can gain the theoretical property of restricted classifier consistency under specific conditions. However, this approach has both practical and theoretical limitations. For the practical limitation, it fails to consider differentiating the true label for each instance in the group, which could limit its empirical performance due to the lack of any purified supervision information of individual instances. For the theoretical limitation, it cannot guarantee the \emph{risk consistency}, i.e., the estimator (by aggregate observations) is \emph{biased} to the classification risk (by fully labeled data). Besides, it has a strict restriction on the used loss function due to the log-likelihood, making it not flexible enough when the loss needs to be changed with the dataset in practice.

In this paper, we propose a universal unbiased method for CFAO, which holds an \emph{unbiased estimator of the classification risk}. Although many previous studies have explored unbiased risk estimators (UREs) to solve specific weakly supervised learning problems \cite{ishida2018binary,cao2021learning,feng2021pointwise}, they only focused on a certain learning problem. This limits the usage of existing UREs, and whether there exists a universal URE for various problems of CFAO is still unknown. We for the first time give an affirmative answer to this question. Our proposed universal URE works by weighing the importance of each label for instance in the group, which could guide the classifier to identify the true label for gaining purified supervision of individual instances. Theoretically, the risk consistency of our method can be naturally guaranteed due to its unbiasedness, which also makes our method compatible with arbitrary losses. Extensive experiments on various problems of CFAO demonstrate the superiority of our method.

\section{Preliminary Knowledge}
In this section, we introduce preliminary knowledge of the supervised classification task and the task of classification from aggregate observations.

\subsection{Supervised Classification}
Let $\mathcal{X}\in\mathbb{R}^d$ be the feature space with $d$ dimensions and $\mathcal{Y}\in [k]$ be the label space. We denote by $\bm{x}\in\mathcal{X}$ a feature vector and $y\in \mathcal{Y}$ the ground-truth label of $\bm{x}$. Each example $(\bm{x},y)$ is supposed to be sampled from an underlying distribution with probability density $p(\bm{x},y)$. For the $k$-class classification task, the goal is to train a learning function $f:\mathcal{X}\mapsto\mathbb{R}^k$ that tries to the classification risk defined as
\begin{gather}
\label{class_risk}
R(f)=\mathbb{E}_{p(\bm{x},y)}\big[\mathcal{L}(\bm{x},y; f)\big],
\end{gather}
where $\mathbb{E}_{p(\bm{x},y)}[\cdot]$ denotes the expected value over $p(\bm{x},y)$ and $\mathcal{L}$ denotes a classification loss (e.g., the softmax cross entropy loss: $\mathcal{L}(\bm{x},y; f) = -\log\frac{\exp(f_y(\bm{x}))}{\sum_{j=1}^k\exp(f_j(\bm{x}))}$), where $f_y(\bm{x})$ is the $y$-th element of $f(\bm{x})$. The predicted label $\hat{y}$ of $\bm{x}$ is given as $\hat{y}=\mathrm{argmax}_{y\in\mathcal{Y}}f_y(\bm{x})$. Since the probability density $p(\bm{x},y)$ is not accessible, we need to collect identically and independently distributed training examples $\{(\bm{x}^{(i)}, y^{(i)})\}_{i=1}^n$ and minimize the empirical version of Eq.~(\ref{class_risk}) instead:
\begin{gather}
\widehat{R}(f)=\sum\nolimits_{i=1}^n\big[\mathcal{L}(\bm{x}^{(i)},y^{(i)}; f)\big],
\end{gather}
which is referred to as \emph{empirical risk minimization} principle~\cite{vapnik1999overview}.

\subsection{Classification from Aggregate Observations}
For the CFAO task, we aim to learn a classifier with only supervision (i.e., aggregate information) on groups of instances. Concretely, given a group of instances $\bm{x}_{1:m}=\{\bm{x}_1,\bm{x}_2\ldots,\bm{x}_m\}$ where $m$ denotes the size of the group, their true labels $y_{1:m}=\{y_1,y_2\ldots,y_m\}$ are unavailable, and we only know an aggregate label $z\in\mathcal{Z}$ where $\mathcal{Z}$ represents the space of aggregate labels. Each aggregate label can be obtained from $y_{1:m}$ via an aggregate function $g:\mathcal{Y}^m \to \mathcal{Z}$, i.e., $z=g(y_{1:m})$. Our goal is to use observations of $(\bm{x}_{1:m},z)$ to train a classifier that can predict the true label as accurately as possible.

In Table \ref{Examples}, we show some typical examples of the CFAO task and their corresponding aggregate functions. All the problems listed in Table \ref{Examples} can be solved by our proposed universal unbiased risk estimator. In what follows, we introduce two necessary assumptions for mathematically formulating the CFAO task, which were also adopted by the previous study \cite{zhang2020learning}.
\begin{assumption}
\label{aggregate_assump}
$p(z|\bm{x}_{1:m},y_{1:m}) = p(z|y_{1:m})$.
\end{assumption}
This assumption indicates that given the true labels $y_{1:m}$, the aggregate label $z$ is independent of the instances $\bm{x}_{1:m}$. This assumption was commonly used in previous studies \cite{carbonneau2018multiple,cui2020classification,zhang2020learning,bao2022pairwise}. It can be implied in the data collection process, e.g., when we first collect $(\bm{x},y)$-pairs but only disclose some summary statistics of $y_{1:m}$ for learning due to privacy concerns. It also means that we
expect the true label $y$ to carry enough information about $\bm{x}$, so that we do not need to extract more
information from $\bm{x}_{1:m}$ to obtain $z$.

\begin{assumption}
\label{independent_assump}
$p(y_{1:m}|\bm{x}_{1:m}) = \prod_{i=1}^m p(y_i|\bm{x}_i)$.
\end{assumption}
This assumption indicates that the examples in the same group are independent. It is worth noting that all the examples in the training set are independently collected, and we extend such an independent property to the group level.

Combing Assumption~\ref{aggregate_assump} and Assumption~\ref{independent_assump}, we can decompose the joint distribution $p(\bm{x}_{1:m}, y_{1:m}, z)$ as
\begin{gather}
\label{joint_decomp}
p(\bm{x}_{1:m},y_{1:m},z) = p(z|y_{1:m})\prod\nolimits_{i=1}^m p(y_i|\bm{x}_i)p(\bm{x}_i),
\end{gather}
which would be beneficial for us to derive a universal unbiased risk estimator. 

\section{The Proposed Method}
In this section, we present our universal unbiased method for CFAO. Throughout this paper, we use a slightly different definition of ``unbiased", i.e., we say that a method is unbiased if the derived risk of this method on weakly supervised data is equivalent to the ordinary classification risk on fully supervised data (as shown in Eq.~(\ref{class_risk})). Hence, our key idea is to solve the problem by risk rewriting \cite{sugiyama2022machine}, i.e., rewriting the classification risk into an equivalent form that can be estimated from the given aggregate observations. Existing risk rewriting methods can only solve a single problem of CFAO. In this paper, we propose a universal method for CFAO, which is presented as follows.

\subsection{Unbiased Risk Estimator}
\begin{theorem}
\label{ure}
The classification risk $R(f)$ in Eq.~(\ref{class_risk}) can be equivalently expressed as follows:
\begin{gather}
\label{expected_risk}
R_{\mathrm{agg}}(f) = \mathbb{E}_{p(\bm{x}_{1:m}, z)}\big[\mathcal{L}_{\mathrm{agg}}(\bm{x}_{1:m}, z; f)\big].
\end{gather}
where $\mathcal{L}_{\mathrm{agg}}(\bm{x}_{1:m}, z; f)$ (with $z=g(y_{1:m})$ and $y_{1:m}$ is unknown to the learning algorithm) is defined as 
\begin{align}
\label{L_agg}
&\mathcal{L}_{\mathrm{agg}}(\bm{x}_{1:m}, z; f) = \frac{1}{m}\frac{1}{p(z|\bm{x}_{1:m})} \cdot\\
\nonumber
&\qquad\qquad \sum\nolimits_{i=1}^{m}\sum\nolimits_{j=1}^{k}p(z,y_i=j | \bm{x}_{1:m})\cdot\mathcal{L}(\bm{x}_i,j;f),
\end{align}
where $\mathcal{L}(\bm{x}_i,j;f)$ is an ordinary classification loss function as discussed in Eq.~(\ref{class_risk}).
\end{theorem}
The proof of Theorem \ref{ure} is provided in Appendix \ref{Appendix_A}.

Theorem \ref{ure} indicates that we could recover the classification risk by using aggregate observations with a specially defined loss function $\mathcal{L}_{\mathrm{agg}}(\bm{x}_{1:m},z;f)$ in Eq.~(\ref{L_agg}). For the introduced loss function $\mathcal{L}_{\mathrm{agg}}$, we do not impose any restrictions on the ordinary classification loss $\mathcal{L}$, hence our method can be compatible with arbitrary losses.
We can observe that $\mathcal{L}_{\mathrm{agg}}$ works as an importance-weighting loss, due to the instance-level probability $p(z,y_i=j | \bm{x}_{1:m})$ and the group-level probability $p(z | \bm{x}_{1:m})$. Specifically, when $p(z,y_i=j | \bm{x}_{1:m})$ is small, the probability of the label $j$ being the true label of $i$-th instance $\bm{x}_i$ in the group $\bm{x}_{1:m}$ is small. This means that $(\bm{x}_i,j)$ is unlikely to be true (i.e., drawn from the fully supervised data distribution $p(\bm{x},y)$), therefore we should assign a small weight to $\mathcal{L}(\bm{x}_i,j;f)$. The group-level probability $p(z|\bm{x}_{1:m})$ can be regarded as a normalization factor that normalizes the instance-level probability $p(z,y_i=j | \bm{x}_{1:m})$.

It is worth noting that in Theorem \ref{ure}, we do not impose any restrictions on the group size $m$, which means that with aggregate observations of arbitrary group sizes, we could still recover the classification risk in Eq.~(\ref{class_risk}).
Therefore, for classification from a given set of aggregate observations $\{(\bm{x}_{1:m}^{(i)},z^{(i)})\}_{i=1}^n$, we can minimize the following empirical risk (i.e., the empirical version of Eq.~(\ref{expected_risk})):
\begin{gather}
\label{empirical_risk}
\widehat{R}_{\mathrm{agg}}(f)=\frac{1}{n}\sum\nolimits_{i=1}^n\mathcal{L}_{\mathrm{agg}}(\bm{x}^{(i)}_{1:m},z^{(i)}; f).
\end{gather}
Here, the key challenge becomes how to calculate $\mathcal{L}_{\mathrm{agg}}$ given an aggregate observation $(\bm{x}_{1:m}^{(i)},z^{(i)})$ and a classifier $f$. According to Eq.~(\ref{L_agg}), we can observe that the remaining issue is to empirically estimate $p(z | \bm{x}_{1:m})$ and $p(z,y_i=j | \bm{x}_{1:m})$. Once the two probabilities $p(z | \bm{x}_{1:m})$ and $p(z,y_i=j | \bm{x}_{1:m})$ are estimated by the classifier $f$, we can substitute the estimated values into $\mathcal{L}_{\mathrm{agg}}$ to obtain the loss value for training the classifier $f$. It is also noteworthy that different problems of CFAO have different rules on the aggregate function $g(y_{1:m})=z$. Therefore, $p(z | \bm{x}_{1:m})$ and $p(z,y_i=j | \bm{x}_{1:m})$ can be estimated in different ways in terms of different problems of CFAO.

\subsection{Analysis from the EM Perspective}
Here, we provide a theoretical justification on how the estimated $\frac{p(z,y_i=j\mid \bm{x}_{1:m})}{p(z\mid \bm{x}_{1:m})}$ helps train the classifier and differentiates the ground truth label. 
We show that our method actually maximizes the likelihood $\log(p(z^{(v)},\bm{x}_{1:m}^{(v)};\theta))$ during training when we employ the widely used cross-entropy loss. Let us denote by $\theta$ the parameters of the classifier, $p(\cdot;\theta)$ the probability function estimated by the classifier, $S(z)=\{y_{1:m}\in \mathcal{Y}^m\mid g(y_{1:m})=z\}$ where $g$ denotes the aggregation function, and $\omega^{(v)}_{y_{1:m}}$ the weight corresponding to $y_{1:m}$ for the $v$-th example in the dataset, where  $0\leq\omega^{(v)}_{y_{1:m}}\leq1$ and $\sum_{y_{1:m}\in S(z^{(v)})}\omega^{(v)}_{y_{1:m}}=1$.
Then, we have the following theorem.
\begin{theorem}
\label{EM_analysis}
The following inequality holds:
\begin{align}
\label{lower_bound}
\log(p(z^{(v)}\mid \bm{x}^{(v)}_{1:m};\theta))\geq \sum\nolimits_{y_{1:m}^{(v)}\in S(z^{(v)})} \omega^{(v)}_{y_{1:m}}&\\
\nonumber
\log(p(y_{1:m}^{(v)},\bm{x}_{1:m}^{(v)};\theta)&/\omega^{(v)}_{y_{1:m}}),
\end{align}
where the inequality holds with equality when $\omega^{(v)}_{y_{1:m}}=p(y_{1:m}^{(v)}\mid \bm{x}_{1:m}^{(v)};\theta)/p(z^{(v)}\mid \bm{x}_{1:m}^{(v)};\theta)$ and Eq.~(\ref{lower_bound}) can be considered identical to Eq.~(\ref{L_agg}) during the training.

\end{theorem}
The proof of Theorem~\ref{EM_analysis} is provided in Appendix~\ref{Appendix_B}.

Thanks to Theorem~\ref{EM_analysis}, our method can be considered as an EM algorithm \cite{EM_algorithm} that maximizes a log-likelihood objective.
At the E-step, our method assigns  $\omega^{(v)}_{y_{1:m}}=\frac{ p(y_{1:m}^{(v)}\mid \bm{x}_{1:m}^{(v)};\theta)}{p(z^{(v)}\mid \bm{x}_{1:m}^{(v)};\theta)}$, to make the inequality holds with equality, i.e., to maximize $\sum_{v=1}^n\sum_{y_{1:m}^{(v)}\in S(z^{(v)})} \omega^{(v)}_{y_{1:m}}\log(\frac{ p(y_{1:m}^{(v)},\bm{x}_{1:m}^{(v)};\theta)}{\omega^{(v)}_{y_{1:m}}})$ with respect to $\omega$ when $p(y_{1:m}^{(v)},\bm{x}_{1:m}^{(v)};\theta)$ is fixed. At the M-step, our method aims to maximize Eq.~(\ref{lower_bound}) with respect to $\theta$ when $\omega$ is fixed (i.e., to train the classifier by maximizing the improved lower bound). 
\section{Realizations of Our Proposed Method}
\label{Realizations}
In this section, we describe the realizations of our method for various problems of CFAO, including classification via pairwise similarity \cite{hsu2019multi}, classification via triplet comparison \cite{zhang2020learning}, multiple-instance learning \cite{carbonneau2018multiple}, and learning from label proportions \cite{yu2013proptosvm}. We also provide the realization of our method for ordinal classification with only ranking or triplet comparison observations, in Appendix \ref{Appendix_C}.

It is worth noting that for classification via pairwise similarity or triplet comparison, the learned classifier is \emph{not identifiable}, which means the identifiable mapping from the classifier outputs to semantic classes is lost. This problem is caused by the extremely limited supervision information provided by these two problems. In this case, the classifier consistency cannot be strictly guaranteed \cite{zhang2020learning} while the risk consistency of our proposed method still holds, which also confirms the superiority of our method.
Pratically, an identifiable mapping could be obtained if we could obtain a permutation of classes in accordance with the learned classifier \cite{zhang2020learning}. A common method to obtain an identifiable mapping is by using a validation dataset to solve the mapping problem. \citet{hsu2019multi} proposed a method to obtain an optimal mapping by solving a linear sum assignment problem using the Hungarian algorithm \cite{kuhn1955hungarian}. In the experiments, we follow \cite{hsu2019multi} to evaluate the performance of classification methods with the obtained optimal mapping.

In what follows, we demonstrate how to estimate $p(z | \bm{x}_{1:m})$ and $p(z,y_i=j | \bm{x}_{1:m})$ of our proposed $\mathcal{L}_{\mathrm{agg}}$ in Eq.~(\ref{L_agg}), for a certain problem of CFAO with the classifier $f$. Once the two probabilities $p(z | \bm{x}_{1:m})$ and $p(z,y_i=j | \bm{x}_{1:m})$ are estimated by the classifier $f$, we can substitute the estimated values into $\mathcal{L}_{\mathrm{agg}}$ to obtain the loss value for training the classifier $f$.

For convenience, we represent $p(y_i=j| \bm{x}_i)$ by $\eta_j(\bm{x}_i)$, which denotes the probability of the true label $y_i$ being the $j$-th class, for the instance $\bm{x}_i$. Given a classifier $f$, $\eta_j(\bm{x}_i)$ can be approximated by applying the softmax function to the classifier output $f(\bm{x}_i)\in\mathbb{R}^k$, i.e., 
\begin{gather}
\label{conf_approx}
\eta_j(\bm{x}_i)=\frac{\exp(f_j(\bm{x}_i))}{\sum_{v=1}^k\exp(f_v(\bm{x}_i))}.
\end{gather}
In what follows, we will demonstrate how $p(z | \bm{x}_{1:m})$ and $p(z,y_i=j | \bm{x}_{1:m})$ of $\mathcal{L}_{\mathrm{agg}}$ can be derived from $\eta_j(\bm{x}_i)$ for various problems of CFAO.

\subsection{Classification via Pairwise Similarity}
In the problem of classification via pairwise similarity, each group has two instances (i.e., $m=2$), and the aggregate information
is whether two instances in the group belong to the same
class (similar) or not (dissimilar). In this case, the aggregation function $g$ is defined as $g(y_{1:2})=\mathbb{I}[y_1=y_2]$ where $\mathbb{I}$ is the indicator function, which returns $1$ if $y_1=y_2$ and $0$ otherwise. This problem was investigated as semi-supervised clustering in early studies \cite{bilenko2004integrating,basu2004probabilistic}. In recent years, this problem has been studied from the perspective of classification \cite{hsu2019multi,zhang2020learning}, based on the maximum likelihood principle \cite{nishii1989maximum}.

We also study this problem from the perspective of classification, and our proposed loss function $\mathcal{L}_{\mathrm{agg}}$ can be applied to solve this problem, by using the following proposition.
\begin{proposition}
For the problem of classification via pairwise similarity ($m=2$), $p(z| \bm{x}_{1:m})$ and $p(z,y_i=j| \bm{x}_{1:m})$ of $\mathcal{L}_{\mathrm{agg}}$ in Eq.~(\ref{L_agg}) can be empirically estimated by 
\begin{align}
\nonumber
p(z=1|\bm{x}_{1:2}) &= \sum\nolimits_{j=1}^k \eta_j(\bm{x}_1)\eta_j(\bm{x}_2),\\
\nonumber
p(z=0|\bm{x}_{1:2}) &= 1 - p(z=1|\bm{x}_{1:2}),
\end{align}
and 
\begin{align}
\nonumber
p(z=1,y_1=j| \bm{x}_{1:2}) &= \eta_j(\bm{x}_1)\eta_j(\bm{x}_2),\\
\nonumber
p(z=1,y_2=j| \bm{x}_{1:2}) &= \eta_j(\bm{x}_1)\eta_j(\bm{x}_2),\\
\nonumber
p(z=0,y_1=j| \bm{x}_{1:2}) &= \left(1-\eta_j(\bm{x}_2)\right)\eta_j(\bm{x}_1),\\
\nonumber
p(z=0,y_2=j| \bm{x}_{1:2}) &= \left(1-\eta_j(\bm{x}_1)\right)\eta_j(\bm{x}_2).
\end{align}
\end{proposition}

\subsection{Classification via Triplet Comparison}
In the problem of classification via triplet comparison, each group has three instances (i.e., $m=3$), and the aggregation information is whether one instance is more similar to the other one, compared with the third one. In this case, the aggregate function $g$ is defined as $g(y_{1:3})=\mathbb{I}\left[d(y_1,y_2)<d(y_1,y_3)\right]$ where $d:\mathcal{Y}\times\mathcal{Y}\mapsto\mathbb{R}$ is a distance measure between classes (a smaller distance means a larger similarity). In our studied classification setting, the distance measure is defined as $d(y,y^\prime)=\mathbb{I}[y\neq y^\prime]$.

Triplet comparison data has been widely studied in metric learning \cite{schultz2003learning,kumar2008semisupervised,sohn2016improved,mojsilovic2019relative}. Recently, \citet{cui2020classification} showed that we can successfully learn a binary classifier from only triplet comparison data, and \citet{zhang2020learning} studied multi-class classification from triplet comparison data. We also study multi-class classification from triplet comparison data, and our proposed loss function $\mathcal{L}_{\mathrm{agg}}$ can be applied to solve this problem, by using the following proposition.
\begin{proposition}
For the problem of classification via triplet comparison ($m=3$), $p(z| \bm{x}_{1:m})$ and $p(z,y_i=j| \bm{x}_{1:m})$ of $\mathcal{L}_{\mathrm{agg}}$ in Eq.~(\ref{L_agg}) can be empirically estimated by
\begin{small}
\begin{align}
\nonumber
p(z=1|\bm{x}_{1:3}) &= \sum\nolimits_{j=1}^k\eta_j(\bm{x}_1)\eta_j(\bm{x}_2)\big(1-\eta_j(\bm{x}_3)\big),\\
\nonumber
p(z=0|\bm{x}_{1:3}) &= 1 - p(z=1|\bm{x}_{1:3}),
\end{align}
\end{small}and 
\begin{small}
\begin{align}
\nonumber
p(z=1,y_1=j| \bm{x}_{1:3}) 
&= \eta_j(\bm{x}_2)\left(1-\eta_j(\bm{x}_3)\right)\eta_j(\bm{x}_1),\\
\nonumber
p(z=1,y_2=j| \bm{x}_{1:3}) 
&= \eta_j(\bm{x}_1)\left(1-\eta_j(\bm{x}_3)\right)\eta_j(\bm{x}_2),\\
\nonumber
p(z=1,y_3=j| \bm{x}_{1:3}) 
&= \sum\nolimits_{v\neq j}\eta_v(\bm{x}_1)\cdot \eta_v(\bm{x}_2)\eta_j(\bm{x}_3),\\
\nonumber
p(z=0,y_1=j| \bm{x}_{1:3}) 
&= \left(1-\eta_j(\bm{x}_2)\left(1-\eta_j(\bm{x}_3)\right)\right)\eta_j(\bm{x}_1),\\
\nonumber
p(z=0,y_2=j| \bm{x}_{1:3}) 
&= \left(1-\eta_j(\bm{x}_1)\left(1-\eta_j(\bm{x}_3)\right)\right)\eta_j(\bm{x}_2),\\
\nonumber
p(z=0,y_3=j| \bm{x}_{1:3}) 
&= \Big(1-\sum\limits_{v\neq j}\eta_v(\bm{x}_1)\cdot \eta_v(\bm{x}_2)\Big)\eta_j(\bm{x}_3).
\end{align}
\end{small}
\end{proposition}

\subsection{Learning from Label Proportions}
Learning from label proportions \cite{yu2014llp,quadrianto2008estimating,ROT,scott2020learning} is also an attractive problem of CFAO.
In learning from label proportions, each group has at least two instances (i.e., $m\geq 2$), and the aggregate information is the proportion of data from each class in the group. In this problem, the aggregate label space $\mathcal{Z}$ is a $k$-dimensional vector space (i.e., $\mathcal{Z}=\mathbb{R}^k$). The aggregate function is defined as
$g_j(y_{1:m})=\sum_{i=1}^{m}\mathbb{I}[y_i=j]=z_j$, which corresponds to the number of instances from each class in the group. It is noteworthy that this aggregate function is slightly different from the original aggregate function mentioned in Table~\ref{Examples}, where the denominator $m$ is removed for realization convenience, which would not affect the natural property of this problem.

We can also solve this problem using our proposed $\mathcal{L}_{\mathrm{agg}}$ by the following proposition.
\begin{proposition}
\label{llp_proposition}
For the problem of learning from label proportions ($m\geq 2$), $p(\bm{z}| \bm{x}_{1:m})$ and $p(\bm{z},y_i=j| \bm{x}_{1:m})$ of $\mathcal{L}_{\mathrm{agg}}$ in Eq.~(\ref{L_agg}) can be empirically estimated by
\begin{gather}
\nonumber
p(\bm{z}| \bm{x}_{1:m})=\sum\nolimits_{y_{1:m}\in\delta(\bm{z})}\prod\nolimits_{i=1}^m\eta_{y_i}(\bm{x}_i),
\end{gather}
where $\delta(\bm{z})=\{y_{1:m} | \bm{g}(y_{1:m})=\bm{z}\}$ and
\begin{align}
\nonumber
&p(\bm{z}=\bm{g}(y_{1:m}),y_i=j| \bm{x}_{1:m})\\
\nonumber
&\qquad\qquad\qquad=\sum\nolimits_{y_{1:m\backslash i}\in \delta(\bm{z},i,j)}\prod\nolimits_{v\neq i}
\eta_{y_v}(\bm{x}_v)\eta_j(\bm{x}_i),
\end{align}
where $\delta (\bm{z},i,j)=\{y_{1:m \backslash i} | \bm{g}(y_{1:m})=\bm{z},y_i=j\}$.
\end{proposition}

\subsection{Multiple-Instance Learning}

\begin{table*}[!t]
\centering
\caption{Statistics of the used benchmark-simulated datasets and the corresponding models. \#Sampled groups represents the number of groups sampled for classification via pairwise similarity/triplet comparison/label proportion.}
\label{dataset}
\resizebox{1.0\textwidth}{!}{
\setlength{\tabcolsep}{2.5mm}{
\begin{tabular}{cccccccc}
\toprule
Dataset         & \#Train & \#Val & \#Test & \#Classes & \#Features & \#Sampled groups & Model                  \\ \midrule
MNIST           & 45,000   & 15,000 & 10,000  & 10        & 784        & 120,000/120,000/30,000     & 5-layer LeNet          \\
Kuzushiji-MNIST & 45,000   & 15,000 & 10,000  & 10        & 784        & 120,000/120,000/30,000     & 5-layer LeNet          \\
Fashion-MNIST   & 45,000   & 15,000 & 10,000  & 10        & 784        & 120,000/120,000/30,000     & 5-layer LeNet          \\
CIFAR-10        & 37,500   & 12,500 & 10,000  & 10        & 3,072      & 120,000/120,000/30,000     & 22-layer DenseNet      \\
SVHN            & 54,942   & 18,315 & 26,032  & 10        & 3,072      & 120,000/120,000/30,000     & 22-layer DenseNet      \\ \midrule
msplice         & 1,905    & 635   & 635    & 3         & 240        & 6,350/6,350/1,587          & 3-layer MLP ($d$-300-$k$) \\
optdigits       & 3,372    & 1,124  & 1,124   & 10        & 62         & 11,240/11,240/2,810        & 3-layer MLP ($d$-300-$k$) \\
pendigits       & 6,594    & 2,199  & 2,199   & 10        & 16         & 21,984/21,984/5,496        & 3-layer MLP ($d$-300-$k$) \\
usps            & 5,578    & 1,860  & 1,860   & 10        & 256        & 18,596/18,596/4,649        & 3-layer MLP ($d$-300-$k$) \\
vehicle         & 507     & 169   & 170    & 4         & 18         & 1,692/1,692/423           & 3-layer MLP ($d$-300-$k$) \\ \bottomrule
\end{tabular}
}
}
\end{table*}

Multiple-instance learning \cite{maron1997framework,zhang2001dd,carbonneau2018multiple,ilse2018attention} is a widely studied weakly supervised learning problem, which also belongs to the task of CFAO. In multiple-instance learning, each group has multiple instances (i.e., $m\geq 2$), and the aggregate information is whether the group has at least one positive instance. Since multiple-instance learning focuses on binary classification (i.e., $k=2$), we define the label space $\mathcal{Y}$ as $\{0,1\}$. In this case, the aggregate function is defined as $g(y_{1:m}) = \max(y_{1:m})$, which means that the group that has at least one positive instance is a positive group, otherwise, it is a negative group. Since the output of the binary classifier $f(\bm{x})\in \mathbb{R}$ is a scalar, we apply the Sigmoid function to approximate $\eta(\bm{x})$, i.e., $\eta_1(\bm{x}_i)=\frac{1}{1+\exp(-f(\bm{x}_i))}$ and $\eta_0(\bm{x}_i)=1-\eta_1(\bm{x}_i)$.

Our proposed loss function $\mathcal{L}_{\mathrm{agg}}$ can also be applied to solve this problem, by the following proposition.
\begin{proposition}
For the problem of multiple-instance learning ($m\geq 2$ and $k=2$), $p(z| \bm{x}_{1:m})$ and $p(z,y_i=j| \bm{x}_{1:m})$ of $\mathcal{L}_{\mathrm{agg}}$ in Eq.~(\ref{L_agg}) can be empirically estimated by
\begin{align}
\nonumber
p(z=0| \bm{x}_{1:m}) &= \prod\nolimits_{i=1}^m \eta_0(\bm{x}_i),\\
\nonumber
p(z=1| \bm{x}_{1:m}) &= 1-p(z=0| \bm{x}_{1:m}),
\end{align}
and 
\begin{align}
\nonumber
p(z=1,y_i=1| \bm{x}_{1:m}) &= \eta_1(\bm{x}_i),\\
\nonumber
p(z=1,y_i=0| \bm{x}_{1:m}) &= (1-\prod\nolimits_{j\neq i}\eta_0(\bm{x}_j))\eta_0(\bm{x}_i),\\
\nonumber
p(z=0,y_i=0| \bm{x}_{1:m}) &=\prod\nolimits_{j\neq i}\eta_0(\bm{x}_j)\eta_0(\bm{x}_i),\\
\nonumber
p(z=0,y_i=1| \bm{x}_{1:m}) &=0.
\end{align}
\end{proposition}

\section{Experiments}
\label{Experiments}
In this section, we conduct extensive experiments to empirically demonstrate the effectiveness of our proposed method in various problems of CFAO. For classification via pairwise similarity/triplet comparison/label proportion, we use five popular large-scale benchmark datasets including MNIST \cite{MNIST}, Kuzushiji-MNIST \cite{KMINST}, Fashion-MNIST \cite{FASHION}, SVHN \cite{SVHN}, and CIFAR-10 \cite{CIFAR10} and five regular-scale datasets from the UCI Machine Learning Repository \cite{Dua2019UCI} including usps, pendidigts, optdigits, msplice, and vehicle. For multiple-instance learning, we use five common benchmark datasets in this area \cite{dietterich1997solving,andrews2002support}, inclduing Musk1, Musk2, Elephant, Fox, and Tiger. Since our proposed method can be compatible with arbitrary models and losses, we use various base models, including 5-layer LeNet \cite{MNIST}, 22-layer DenseNet \cite{Dense}, and 3-layer ($d$-300-$k$) Multilayer Perceptron on the above datasets. For the classification loss $\mathcal{L}$ in our proposed $\mathcal{L}_{\mathrm{agg}}$, we simply adopt the widely used softmax cross entropy loss for multi-class classification and adopt the logistic loss for binary classification. Detailed descriptions of the used datasets and the corresponding models are provided in Table \ref{dataset}. The details of our algorithmic procedure, hyperparameter settings, and the characteristics of datasets for multiple-instance learning are provided in Appendix~\ref{Appendix_Experiments}.

For classification via pairwise similarity/triplet comparison/label proportion, the aggregate observations are randomly generated from the training set with replacement, according to Assumption \ref{independent_assump}. Since the learned classifier for classification via pairwise similarity/triplet comparison is not identifiable, we follow \citet{hsu2019multi,zhang2020learning} to evaluate the performance by modified accuracy that allows any permutation of classes, and the optimal permutation is obtained by solving a linear sum assignment problem using the Hungarian algorithm \cite{kuhn1955hungarian}.

We run five trials on each dataset and record the mean accuracy and  standard deviation (mean $\pm$ std). The best performance among all the methods is highlighted in boldface. We also conduct paired \textit{t}-test at 5\% significance level, and use $\bullet/\circ$ to denote whether our proposed \underline{u}niversal \underline{u}nbiased \underline{m}ethod (UUM) is significantly better/worse than a compared method. 

\begin{table*}[ht]
\caption{Test accuracy (mean $\pm$ std) of each method for classification via pairwise similarity on large-scale benchmark datasets.}
\label{similarity_benchmark}
\centering
\resizebox{1.0\textwidth}{!}{
\setlength{\tabcolsep}{3.5mm}{
\begin{sc}
\begin{tabular}{c|ccccc}
\toprule
 Method & MNIST              & Kuzushiji          & Fashion            & CIFAR10            & SVHN               \\ \midrule
log-likelihood & $98.87\pm 0.06\%\bullet$  & $93.40\% \pm 0.64\%$      & $88.75 \pm 0.31\%\bullet$ & $71.45 \pm 0.47\%\bullet$ & $91.06 \pm 0.24\%\bullet$ \\
Siamese        & $98.63 \pm 0.15\%\bullet$ & $89.25 \pm 1.87\%\bullet$ & $82.20 \pm 1.45\%\bullet$ & $58.16 \pm 1.92\%\bullet$ & $87.42 \pm 4.06\%\bullet$ \\
Contrastive     & $98.86 \pm 0.14\%\bullet$ & $93.63 \pm 0.53\%\bullet$ & $89.17 \pm 0.31\%$ & $23.69 \pm 0.65\%\bullet$ & $91.30 \pm 0.33\%\bullet$ \\  \midrule
UUM (ours)             & $\bm{98.99 \pm 0.04\%}$   & $\bm{94.04 \pm 0.50\%}$   & $\bm{89.19  \pm 0.37\%}$  & $\bm{72.52 \pm 0.68\%}$   & $\bm{92.41 \pm 0.46\%}$   \\
\bottomrule
\end{tabular}
\end{sc}
}
}
\end{table*}
\begin{table*}[ht]
\centering
\caption{Test accuracy (mean $\pm$ std) of each method for classification via pairwise similarity on regular-scale UCI datasets.}
\label{similarity_UCI}
\resizebox{1.0\textwidth}{!}{
\setlength{\tabcolsep}{3.5mm}{
\begin{sc}
\begin{tabular}{c|ccccc}
\toprule
Method      & msplice            & optdigits          & pendigits          & usps               & vehicle            \\ \midrule
log-likelihood & $94.93 \pm 0.43 \%$        & $98.10 \pm 0.49 \%$ & $88.18 \pm 14.14\%$ & $96.87 \pm 0.17 \%\bullet$ & $\bm{79.41 \pm 4.23 \%}$   \\
Siamese         & $94.62 \pm 0.42\% $        & $90.05 \pm 3.50\%\bullet $ & $77.13 \pm 1.95\% \bullet$ & $81.96 \pm 6.16\%\bullet $ & $45.76 \pm 5.29\% $ \\
contrastive     & $93.36 \pm 0.36\% \bullet$ & $95.78 \pm 0.83\% \bullet$ & $90.50 \pm 1.14\% \bullet$ & $93.52 \pm 0.46\% \bullet$ & $60.35 \pm 2.31\%$ \\ \midrule
UUM (ours)            & $\bm{94.99 \pm 0.71 \%}$   & $\bm{98.31 \pm 0.37 \%}$   & $\bm{96.95 \pm 4.14 \%}$   & $\bm{97.02 \pm 0.13 \%}$   & $78.71 \pm 4.20 \%$        \\
\bottomrule
\end{tabular}
\end{sc}
}
}
\end{table*}

\begin{table*}[!t]
\centering
\caption{Test accuracy (mean $\pm$ std) of each method for classification via triplet comparison on large-scale benchmark datasets.}
\label{triplet_benchmark}
\resizebox{1.0\textwidth}{!}{
\setlength{\tabcolsep}{3.5mm}{
\begin{sc}
\begin{tabular}{c|ccccc}
\toprule
Method & MNIST & Kuzushiji & Fashion & CIFAR10 & SVHN \\ \midrule
log-likelihood & $98.92\pm 0.11\%$        & $93.46 \pm 0.10\%\bullet$       & $88.76 \pm 0.28\%$ & $70.57 \pm 0.59\%\bullet$ & $90.61 \pm 0.60\%\bullet$ \\
Triplet        & $97.40\pm 0.10\%\bullet$ & $85.65\pm 0.98\%\bullet$ & $82.14\pm 2.68\%\bullet$  & $47.09 \pm 3.80\%\bullet$ & $84.10 \pm 2.15\%\bullet$ \\
(2+1) Tuple    & $96.85\pm 0.33\%\bullet$ & $80.26\pm 1.46\%\bullet$ & $74.19\pm 1.12\%\bullet$  & $45.53\pm 1.58\%\bullet$  & $74.02 \pm 1.28\%\bullet$ \\ \midrule
UUM (ours)       & $\bm{99.06 \pm 0.09\%} $ & $\bm{93.88 \pm 0.29\%}$  & $\bm{89.18  \pm 0.47\%}$  & $\bm{72.41 \pm 0.98\%}$   & $\bm{92.44 \pm 0.29\%}$   \\\bottomrule
\end{tabular}
\end{sc}
}
}
\end{table*}
\begin{table*}[!t]
\centering
\caption{Test accuracy (mean $\pm$ std) of each method for classification via triplet comparison on regular-scale UCI datasets.}
\label{triplet_UCI}
\resizebox{1.0\textwidth}{!}{
\setlength{\tabcolsep}{3.5mm}{
\begin{sc}
\begin{tabular}{l|lllll}
\toprule
Dataset    & msplice                  & optdigits                & pendigits                 & usps                     & vehicle                  \\ \midrule
log-likelihood & $95.09 \pm 0.90 \%$        & $\bm{98.15 \pm 0.21 \%}$   & $64.30 \pm 35.63 \%\bullet$ & $96.42 \pm 0.34 \%$        & $75.53 \pm 3.28 \%$ \\
triplet        & $92.69 \pm 1.47\%\bullet $ & $91.81 \pm 1.31\%\bullet $ & $65.66 \pm 17.16\% $        & $87.04 \pm 3.46\%\bullet $ & $65.88 \pm 5.37\%\bullet $ \\
(2+1)tuple & $91.43\pm 1.32\%\bullet$ & $81.64\pm 6.29\%\bullet$ & $64.36\pm 0.12\%$         & $74.72\pm 0.96\%\bullet$ & $63.88\pm 3.50\%\bullet$ \\ \midrule
UUM (ours)       & $\bm{95.37 \pm 1.10 \%}$ & $98.13 \pm 0.41 \%$ & $\bm{66.48 \pm 36.24 \%}$ & $\bm{96.49 \pm 0.59 \%}$ & $\bm{76.71 \pm 2.06 \%}$ \\
\bottomrule
\end{tabular}
\end{sc}
}
}
\end{table*}

\subsection{Classification via Pairwise Similarity}
\noindent\textbf{Experimental setup.}
For classification via pairwise similarity, the size of the generated training set is 120,000 for large-scale benchmark datasets and is twice the size of the original training set for regular-scale UCI datasets. We compare our proposed method with three methods of classification via pairwise similarity, including the universal method based on log-likelihood \cite{zhang2020learning} and two representation/metric learning methods including the Siamese network \cite{siamese} and the contrastive loss \cite{contrastive}. Since the output of the two representation/metric learning methods is a vector representation, we use the $K$-means clustering algorithm \cite{bock2007clustering} on vector representations to obtain unidentifiable class predictions and evaluate the performance by modified accuracy with the optimal permutation of classes \cite{kuhn1955hungarian}.

\noindent\textbf{Experimental results.}
Table \ref{similarity_benchmark} and Table \ref{similarity_UCI} report the test accuracy (mean $\pm$ std) of each method for classification via pairwise similarity on large-scale benchmark datasets and UCI datasets, respectively. As can be observed from Table \ref{similarity_benchmark}, UUM achieves the best performance among all the methods on all the large-scale benchmark datasets and significantly outperforms the compared methods in most cases. Table \ref{similarity_UCI} shows that UUM achieves the best performance on 4 out of 5 UCI datasets. From the two tables, we can see that UUM significantly outperforms other compared methods when a complex model (i.e., DenseNet) is used on large-scale datasets. This phenomenon indicates that UUM could obtain a more precise estimation of the importance of each label for each instance in the group, with a powerful model on larger-scale datasets.

\begin{table*}[!t]
\centering
\caption{Test accuracy (mean $\pm$ std) of each method for learning from label proportions on large-scale benchmark datasets.}
\label{LLP_benchmark}
\resizebox{1.0\textwidth}{!}{
\setlength{\tabcolsep}{3.5mm}{
\begin{sc}
\begin{tabular}{llllll}
\toprule
Dataset & MNIST & Kuzushiji & Fashion & CIFAR10 & SVHN \\ \midrule
log-likelihood & $\bm{98.99\pm 0.06\%}$ & $94.06\% \pm 0.27\%\bullet$ & $89.72 \pm 0.23\%$        & $70.56 \pm 0.39\%$        & $90.61 \pm 0.43\%$      \\
ROT            & $98.95 \pm 0.08\% $    & $94.32\% \pm 0.22\%$        & $89.40 \pm 0.31\%\bullet$ & $70.44 \pm 0.48\%\bullet$ & $\bm{92.09 \pm 0.17\%}$\\ \midrule
UUM (ours)        & $98.96 \pm 0.10\% $    & $\bm{94.41 \pm 0.16\%}$     & $\bm{89.87  \pm 0.20\%}$  & $\bm{70.77 \pm 0.66\%}$   & $91.76 \pm 0.18\%$      \\
\bottomrule
\end{tabular}
\end{sc}
}
}
\end{table*}
\begin{table*}[!t]
\centering
\caption{Test accuracy (mean $\pm$ std) of each method for learning from label proportions on regular-scale UCI datasets.}
\label{LLP_UCI}
\resizebox{1.0\textwidth}{!}{
\setlength{\tabcolsep}{3.5mm}{
\begin{sc}
\begin{tabular}{cccccc}
\toprule
Dataset & msplice & optdigits & pendigits & usps & vehicle \\ \midrule
log-likelihood & $95.12 \pm 0.84 \%\bullet$ & $98.35 \pm 0.38 \%$      & $99.35 \pm 0.13 \%$      & $96.89 \pm 0.31 \%\bullet$ & $79.39 \pm 1.44 \%$      \\
ROT            & $95.21 \pm 0.79\% \bullet$ & $98.40 \pm 0.27\% $      & $99.35 \pm 0.17\% $      & $97.18 \pm 0.32\% $        & $\bm{80.14 \pm 0.23\%} $\\ \midrule
UUM (ours)      & $\bm{95.62 \pm 0.64 \%}$   & $\bm{98.43 \pm 0.38 \%}$ & $\bm{99.38 \pm 0.17 \%}$ & $\bm{97.24 \pm 0.41 \%}$   & $79.41 \pm 1.93 \%$      \\ \bottomrule
\end{tabular}
\end{sc}
}
}
\end{table*}

\begin{table*}[!t]
\caption{Test accuracy (mean $\pm$ std) of each method for multiple-instance learning on common benchmark datasets.}
\label{mil_table}
\resizebox{1.0\textwidth}{!}{
\setlength{\tabcolsep}{3.5mm}{
\begin{small}
\begin{sc}
\begin{tabular}{cccccc}
\toprule
Dataset        & elephant               & fox                    & tiger                  & musk1                    & musk2                     \\ \midrule
log-likelihood & $98.50\pm 1.27\%$      & $94.30\pm 1.96\%$      & $98.70\pm 0.67\%$      & $99.35\pm 1.46\%\bullet$ & $80.20\pm 12.70\%\bullet$ \\
minimax-feature & $90.90\pm 2.46\%\bullet$ & $81.00\pm 2.65\%\bullet$ & $91.70\pm 1.72\%\bullet$ & $98.59\pm 1.46\%\bullet$ & $98.43\pm 1.29\%\bullet$ \\ \midrule
UUM (ours)      & $\bm{99.00\pm 1.06\%}$ & $\bm{94.90\pm 1.98\%}$ & $\bm{99.00\pm 0.61\%}$ & $\bm{100.00\pm 0.00\%}$  & $\bm{99.61\pm 0.37\%}$    \\\bottomrule
\end{tabular}
\end{sc}
\end{small}
}
}
\end{table*}
\subsection{Classification via Triplet Comparison}
\noindent\textbf{Experimental setup.}
For classification via triplet comparison, the size of the generated training set is 120,000 for large-scale benchmark datasets and twice the size of the original training set for UCI datasets. 
We compare our proposed UUM with three methods of classification via triplet comparison, including the universal method based on log-likelihood \cite{zhang2020learning} and two representation/metric learning methods including the triplet loss \cite{tripletloss} and the (2+1) tuple loss \cite{sohn2016improved}. We also use the $K$-means clustering algorithm \cite{bock2007clustering} on vector representations obtained by the triplet loss and the (2+1) tuple loss to get unidentifiable class predictions and evaluate the performance by modified accuracy with the optimal permutation of classes \cite{kuhn1955hungarian}.

\noindent\textbf{Experimental results.}
Table \ref{triplet_benchmark} and Table \ref{triplet_UCI} report the test accuracy (mean $\pm$ std ) of each method for classification via triplet comparison on large-scale benchmark datasets and UCI datasets, respectively. The experiments for classification via triplet comparison show similar results compared with the experiments for classification via pairwise similarity. Hence the superiority of our proposed method is also demonstrated. The experimental results for classification via triplet comparison also demonstrate that the performance of UUM would be more remarkable when large-scale datasets and more complex models are used.

\subsection{Learning from Label Proportions}
\noindent\textbf{Experimental setup.}
For learning from label proportions, the size of the generated training set is 30,000 for benchmark datasets and half of the size of the original training set for UCI datasets. The group size $m$ is set to 6. We compare our proposed method with two methods of learning from label proportions, including the log-likelihood method \cite{zhang2020learning} and ROT \cite{ROT}.
\noindent\textbf{Experimental results.}
Table \ref{LLP_benchmark} and Table \ref{LLP_UCI} report the test accuracy of each method for learning from label proportions on large-scale datasets and UCI datasets, respectively. We can find that UUM achieves the best performance for learning from label proportions in most cases. Hence the effectiveness of our proposed method is also demonstrated in the problem of learning from label proportions.



\subsection{Multiple-Instance Learning}
\noindent\textbf{Experimental setup.}
For multiple-instance learning, we collect 5 widely used benchmark datasets, including Elephant, Fox, Tiger, Musk1, and Musk2. We randomly split the given datasets into training, validation, and test sets by 60\%, 20\% and 20\% for each trial. Since these datasets only contain aggregate labels, we evaluate the performance by group-level accuracy. 
We compare our UUM with two methods, including the log-likelihood method \cite{zhang2020learning} and the minimax-feature method \cite{gartner2002multi}.  We use the linear model as the base model to realize our method and the compared methods, for fair comparison.

\noindent\textbf{Experimental results.}
Table \ref{mil_table} reports the test accuracy of each method for multiple-instance learning on common benchmark datasets. It can be seen that our proposed method achieves the best performance on all datasets and significantly outperforms the minimax-feature method in all cases. Therefore, the effectiveness of our method in multiple-instance learning is also validated.

\section{Conclusion}
In this paper, we investigated an interesting learning task called classification from aggregate observations, where we aim to learn a classifier with supervision on groups of instances, instead of supervision on individual instances. This task is quite general and contains a variety of learning problems such as multiple-instance learning and learning from label proportions. To handle this task, we proposed a novel universal method that holds an unbiased estimator of the classification risk for arbitrary losses. Our method has both practical and theoretical advantages. Practically, our method works by importance weighting for each instance and each label in the group, which provides purified supervision for the classifier to learn. Theoretically, our provided unbiased risk estimator not only guarantees the risk consistency of our method but also can be compatible with arbitrary losses. Comprehensive experimental results validated the effectiveness of our proposed method in various problems of classification from aggregate observations. In future work, we plan to extend our proposed universal unbiased method to the regression setting with aggregate observations.

\section*{Acknowledgements}
This research is supported, in part, by the Joint NTU-WeBank Research Centre on Fintech (Award No: NWJ-2021-005), Nanyang Technological University, Singapore. Lei Feng is also supported by the National Natural Science Foundation of China (Grant No. 62106028), Chongqing Overseas Chinese Entrepreneurship and Innovation Support Program, CAAI-Huawei MindSpore Open Fund, and Chongqing Artificial Intelligence Innovation Center. Bo Han is supported by NSFC Young Scientists Fund No. 62006202, and Guangdong Basic and Applied Basic Research Foundation No. 2022A1515011652.
\bibliography{camera_ready}
\bibliographystyle{icml2023}

\newpage
\appendix
\onecolumn

\section{Proof of Theorem \ref{ure}}\label{Appendix_A}
The classification risk $R(f)=\mathbb{E}_{p(\bm{x},y)}[\mathcal{L}(\bm{x},y; f)]$ can be expressed as
\begin{gather}
\nonumber
R(f)=\int_{\mathcal{X}}\sum\nolimits_{y=1}^k p(\bm{x},y)\mathcal{L}(\bm{x},y; f)\mathrm{d}\bm{x}.
\end{gather}
When we take $m$ examples into one group, we can represent $R(f)$ as follows:
\begin{align}
\nonumber
R(f)&=\int_{\mathcal{X}}\sum_{y_{1:m}}p(\bm{x}_{1:m},y_{1:m})\frac{1}{m}\sum_{i=1}^m\mathcal{L}(\bm{x}_i,y_i; f)\mathrm{d}\bm{x}_{1:m}\\
\nonumber
&=\int_{\mathcal{X}}p(\bm{x}_{1:m})\sum_{y_{1:m}}p(y_{1:m}|\bm{x}_{1:m})\frac{1}{m}\sum_{i=1}^m\mathcal{L}(\bm{x}_i,y_i; f)\mathrm{d}\bm{x}_{1:m}\\
\nonumber
&=\mathbb{E}_{p(\bm{x}_{1:m})}\Big[\sum\limits_{y_1:m}p(y_{1:m}|\bm{x}_{1:m})\frac{1}{m}\sum\limits_{i=1}^m\mathcal{L}(\bm{x}_i,y_i; f)\Big]\\
\nonumber
&=\mathbb{E}_{p(\bm{x}_{1:m})}\Big[\sum\nolimits_{y_1:m}p(z|\bm{x}_{1:m})\frac{p(y_{1:m}|\bm{x}_{1:m})}{p(z|\bm{x}_{1:m})}\cdot
\frac{1}{m}\sum\nolimits_{i=1}^m\mathcal{L}(\bm{x}_i,y_i; f)\Big]\\
\nonumber
&=\mathbb{E}_{p(\bm{x}_{1:m})}\Big[ \sum_{z\in\mathcal{Z}}p(z|\bm{x}_{1:m})\sum_{y_{1:m}\in S(z)}\frac{p(y_{1:m}|\bm{x}_{1:m})}{p(z|\bm{x}_{1:m})}\cdot
\frac{1}{m}\sum\nolimits_{i=1}^m\mathcal{L}(\bm{x}_i,y_i; f)\Big],
\end{align}
where $S(z)=\{y_{1:m}\in\mathcal{Y}^m|g(y_{1:m})=z\}$ contains all the possibilities of $y_{1:m}$ that satisfy the condition $g(y_{1:m})=z$. Then, we have
\begin{align}
\nonumber
R(f)&=\mathbb{E}_{p(\bm{x}_{1:m})}\Big[ \sum_{z\in\mathcal{Z}}p(z|\bm{x}_{1:m})\sum_{y_{1:m}\in S(z)}\frac{p(y_{1:m}|\bm{x}_{1:m})}{p(z|\bm{x}_{1:m})}\cdot
\frac{1}{m}\sum_{i=1}^m\mathcal{L}(\bm{x}_i,y_i; f)\Big]\\
\label{identical_form}
&=\mathbb{E}_{p(\bm{x}_{1:m}, z)}\Big[\sum_{y_{1:m}\in S(z)}\frac{p(y_{1:m}|\bm{x}_{1:m})}{p(z| \bm{x}_{1:m})}\cdot
\frac{1}{m}\sum_{i=1}^m\mathcal{L}(\bm{x}_i,y_i; f)\Big]\\
\label{tmp_rf}
&=\mathbb{E}_{p(\bm{x}_{1:m},z)}\Big[\frac{1}{m}\frac{1}{p(z|\bm{x}_{1:m})}\sum_{i=1}^{m}\sum_{j=1}^{k}\sum_{y_{1:m\backslash i}\in S(z,j)}
p(y_{1:m\backslash i}| \bm{x}_{1:m\backslash i})\cdot p(y_i=j| \bm{x}_i)\cdot \mathcal{L}(f(\bm{x}_i),j)\Big],
\end{align}
where we have switched the two summations in the last equality above, and $y_{1:m\backslash i} = \{y_1, \ldots, y_{i-1}, y_{i+1}, \ldots, y_m\}$ and
$S(z,j)=\{y_{1:m\backslash i}\in \mathcal{Y}^{m-1}\mid g(y_1\cdots y_{i-1},j,y_{i+1}\cdots,y_m)=z\}$ contains all the possibilities of $y_{1:m\backslash i}$ on the condition of $y_i=j$. It is worth noting that
\begin{align}
\nonumber
&\sum_{y_{1:m\backslash i}\in S(z,j)}p(y_{1:m\backslash i}|\bm{x}_{1:m\backslash i})\\
\nonumber
&=\frac{\sum_{y_{1:m\backslash i}\in S(z,j)}p(y_{1:m\backslash i}| \bm{x}_{1:m\backslash i})p(y_i=j| \bm{x}_i)}{p(y_i=j| \bm{x}_i)}\\
\label{inter_rf}
&=\frac{p(g(y_{1:m})=z,y_i=j| \bm{x}_{1:m})}{p(y_i=j| \bm{x}_i)}\\
&=\frac{p(z,y_i=j| \bm{x}_{1:m})}{p(y_i=j| \bm{x}_i)}
\end{align}
By substituting Eq.~(\ref{inter_rf}) into Eq.~(\ref{tmp_rf}), we obtain
\begin{align}
\nonumber
R(f)&=\mathbb{E}_{p(\bm{x}_{1:m},z)}\Big[\frac{1}{m}\frac{1}{p(z|\bm{x}_{1:m})}\sum_{i=1}^{m}\sum_{j=1}^{k}\sum_{y_{1:m\backslash i}\in S(z,j)}
p(y_{1:m\backslash i}| \bm{x}_{1:m\backslash i})\cdot p(y_i=j| \bm{x}_i)\cdot \mathcal{L}(f(\bm{x}_i),j)\Big]\\
\nonumber
&=\mathbb{E}_{p(\bm{x}_{1:m},z)}\Big[\frac{1}{m}\frac{1}{p(z|\bm{x}_{1:m})}\cdot
\sum_{i=1}^{m}\sum_{j=1}^{k} p(z,y_i=j| \bm{x}_{1:m})\cdot \mathcal{L}(f(\bm{x}_i),j)\Big]\\
\nonumber
&=R_{\mathrm{agg}}(f),
\end{align}
where completes the proof of Theorem \ref{ure}.

\section{Proof of Theorem \ref{EM_analysis}}\label{Appendix_B}
The log likelihood $\sum_{v=1}^n \log(p(z^{(v)},x_{1:m}^{(v)};\theta))$ could be transformed into:

\begin{align}
&\sum_{v=1}^n \log(p(z^{(v)},x_{1:m}^{(v)};\theta))\\
&=\sum_{v=1}^n \log(\sum_{y_{1:m}^{(v)}\in S(z^{(v)})}p(z^{(v)},y_{1:m}^{(v)},x_{1:m}^{(v)};\theta))\\
&=\sum_{v=1}^n \log(\sum_{y_{1:m}^{(v)}\in S(z^{(v)})}p(z^{(v)}\mid y_{1:m}^{(v)},x_{1:m}^{(v)}) p(y_{1:m}^{(v)},x_{1:m}^{(v)};\theta))\\
&=\sum_{v=1}^n \log(\sum_{y_{1:m}^{(v)}\in S(z^{(v)})} p(y_{1:m}^{(v)},x_{1:m}^{(v)};\theta))\\
&=\sum_{v=1}^n \log(\sum_{y_{1:m}^{(v)}\in S(z^{(v)})} \omega^{(v)}_{y_{1:m}}\frac{ p(y_{1:m}^{(v)},x_{1:m}^{(v)};\theta)}{\omega^{(v)}_{y_{1:m}}})\\
\label{obj}
&\geq\sum_{v=1}^n\sum_{y_{1:m}^{(v)}\in S(z^{(v)})} \omega^{(v)}_{y_{1:m}} \log(\frac{ p(y_{1:m}^{(v)},x_{1:m}^{(v)};\theta)}{\omega^{(v)}_{y_{1:m}}}),
\end{align}
where the second equality is due to $p(z|y_{1:m})=p(z|y_{1:m},x_{1:m})=1$ when $g(y_{1:m})=z$.
The last inequality relies on Jensen’s inequality and the properties of the weight $\omega^{(v)}_{y_{1:m}}$ (i.e., $ 0\leq\omega^{(v)}_{y_{1:m}}\leq1$ and $\sum_{y_{1:m}\in S(z^{(v)})}\omega^{(v)}_{y_{1:m}}=1$). 
The inequality holds with equality when $\frac{ p(y_{1:m}^{(v)},x_{1:m}^{(v)};\theta)}{\omega^{(v)}_{y_{1:m}}}$ is a constant. i.e., $\frac{ p(y_{1:m}^{(v)},x_{1:m}^{(v)};\theta)}{\omega^{(v)}_{y_{1:m}}}=C$ where $C$ is a constant. 

In this case, we have
\begin{align}
\frac{ p(y_{1:m}^{(v)},x_{1:m}^{(v)};\theta)}{C}&=\omega^{(v)}_{y_{1:m}}\\
\sum_{y_{1:m}^{(v)}\in S(z^{(v)})}\frac{ p(y_{1:m}^{(v)},x_{1:m}^{(v)};\theta)}{C}&=\sum_{y_{1:m}^{(v)}\in S(z^{(v)})}\omega^{(v)}_{y_{1:m}}\\
\sum_{y_{1:m}^{(v)}\in S(z^{(v)})}\frac{ p(y_{1:m}^{(v)},x_{1:m}^{(v)};\theta)}{C} &= 1\\
\sum_{y_{1:m}^{(v)}\in S(z^{(v)})}p(y_{1:m}^{(v)},x_{1:m}^{(v)};\theta) &= C\\
\sum_{y_{1:m}^{(v)}\in S(z^{(v)})}p(y_{1:m}^{(v)},x_{1:m}^{(v)},z^{(v)};\theta) &= C\\
p(x_{1:m}^{(v)},z^{(v)};\theta) &= C,
\end{align}
where the last derivation is due to the fact that we exhausted all the possible $y_{1:m}^{(v)}$ in $S(z^{(v)})$. In this way, we have
\begin{gather}
\omega^{(v)}_{y_{1:m}}=\frac{ p(y_{1:m}^{(v)},x_{1:m}^{(v)};\theta)}{p(x_{1:m}^{(v)},z^{(v)};\theta)} =
\frac{ p(y_{1:m}^{(v)}\mid x_{1:m}^{(v)};\theta)}{p(z^{(v)}\mid x_{1:m}^{(v)};\theta)}.
\end{gather}
Therefore, the E-step of our method is to set $\omega^{(v)}_{y_{1:m}}=\frac{ p(y_{1:m}^{(v)}\mid x_{1:m}^{(v)};\theta)}{p(z^{(v)}\mid x_{1:m}^{(v)};\theta)}$, to make the inequality holds with equality, i.e., to maximize $\sum_{v=1}^n\sum_{y_{1:m}^{(v)}\in S(z^{(v)})} \omega^{(v)}_{y_{1:m}}\log(\frac{ p(y_{1:m}^{(v)},x_{1:m}^{(v)};\theta)}{\omega^{(v)}_{y_{1:m}}})$ with respect to $\omega$ when $p(y_{1:m}^{(v)},x_{1:m}^{(v)};\theta)$ is fixed.

On the other hand, the M-step of our method is to maximize $\sum_{v=1}^n\sum_{y_{1:m}^{(v)}\in S(z^{(v)})} \omega^{(v)}_{y_{1:m}}\log(\frac{ p(y_{1:m}^{(v)},x_{1:m}^{(v)};\theta)}{\omega^{(v)}_{y_{1:m}}})$ (i.e., Eq.~(\ref{obj})) with respect to $\theta$ when $\omega$ is fixed.

For the M-step, $\omega$ is fixed, thus we have
\begin{align}
&\log(\frac{ p(y_{1:m}^{(v)},x_{1:m}^{(v)};\theta)}{\omega^{(v)}_{y_{1:m}}})\\
&=\log(\frac{ p(y_{1:m}^{(v)},x_{1:m}^{(v)};\theta)}{p(x_{1:m};\theta)})+(\log(p(x_{1:m};\theta))-\log(\omega^{(v)}_{y_{1:m}}))\\
&=\log(p(y_{1:m}|x_{1:m});\theta)+\log(\frac{p(x_{1:m};\theta)}{\omega^{(v)}_{y_{1:m}}})\\
&=\sum_{i=1}^m\log(p(y_i|x_i;\theta))+\log(\frac{p(x_{1:m};\theta)}{\omega^{(v)}_{y_{1:m}}})\\
&=\sum_{i=1}^m\log(p(y_i|x_i;\theta))+\log(\frac{p(x_{1:m})}{\omega^{(v)}_{y_{1:m}}}),
\end{align}
where the last term $\log(\frac{p(x_{1:m})}{\omega^{(v)}_{y_{1:m}}})$ is a constant when $\omega$ is fixed. 
When the cross-entropy loss is applied, $\mathcal{L}(x_i,y_i;f)=-\log(p(y_i|x_i;\theta))$. Therefore, maximizing $\log(p(y_{1:m}|x_{1:m}))$ is equivalent to minimizing $\mathcal{L}(x_i,y_i;f)$.

It is noteworthy that the objective function we analyze above (i.e., $\sum_{y_{1:m}^{(v)}\in S(z^{(v)})} \omega^{(v)}_{y_{1:m}} \log(\frac{ p(y_{1:m}^{(v)},x_{1:m}^{(v)};\theta)}{\omega^{(v)}_{y_{1:m}}})$) can be considered identical to the Eq.~(\ref{identical_form}), i.e., $\sum_{y_{1:m}\in S(z^{(v)})} \frac{p(y_{1:m}|x_{1:m})}{p(z|x_{1:m})}\cdot \frac{1}{m}\mathcal{L}(x_i,y_i;f)$ except for the differences of the two constant terms $\frac{1}{m}$ and $\log(\frac{p(x_{1:m})}{\omega^{(v)}_{y_{1:m}}})$ when training the classifier, which is also identical to the final objective function used in our paper. 

In summary, the E-step of our method improves the lower bound (i.e., Eq.~(\ref{obj})) of the likelihood $\log(p(z^{(v)},x_{1:m}^{(v)};\theta))$ and the M-step of our method trains the classifier by maximizing the improved lower bound, which indicates that our method actually maximizes the likelihood $\log(p(z^{(v)},x_{1:m}^{(v)};\theta))$.

\section{Additional CFAO Problems}\label{Appendix_C}
\subsection{Rank observation}
In the next two aggregate observation learning tasks, we focus on ordinal regression. Ordinal regression has a similar label space $\mathcal{Y}=\{1,2\cdots k\}$ compared with muti-class classification. But there exists an order between different labels in label space, i.e., $1 < 2 < 3 \cdots <k$. For notation convenience, we use $\eta_j(\bm{x}_i)$ to denote the probability $p(y_i\leq j|\bm{x}_i)$ of the true label $y_i$ less or equal to $j$. Specifically, $\eta_0(\bm{x}_i)=0$ and $\eta_k(\bm{x}_i)=1$, and $p(y_i=j|\bm{x}_i)$ can be calculated by $\eta_j(\bm{x}_i)-\eta_{j-1}(\bm{x}_i)$. In the rank observation task, there are two instances in one bag (i.e., $m=2$), and the aggregate function is defined as $g(y_{1:2})=\mathbb{I}[y_1<y_2]$.

This task is also a CFAO task in an ordinal classification setting, which can be solved by the following proposition.
\begin{proposition}
For the problem of classification via ordinal ranks ($m=2$), $p(z| \bm{x}_{1:m})$ and $p(z,y_i=j| \bm{x}_{1:m})$ of $\mathcal{L}_{\mathrm{agg}}$ in Eq.~(\ref{L_agg}) can be empirically estimated by
\begin{align}
\nonumber
p(z=1| \bm{x}_{1:2}) &= \sum_{j=1}^k \eta_{j-1}(\bm{x}_1)(\eta_j(\bm{x}_2)-\eta_{j-1}(\bm{x}_2)),\\
\nonumber
p(z=0| \bm{x}_{1:2}) &= 1-p(z=1| \bm{x}_{1:2}),
\end{align}
and
\begin{align}
\nonumber
p(z=1,y_1=j| \bm{x}_{1:2})=&\left(1-\eta_j(\bm{x}_2)\right)
(\eta_j(\bm{x}_1)-\eta_{j-1}(\bm{x}_1)),\\
\nonumber
p(z=1,y_2=j| \bm{x}_{1:2})=&\eta_{j-1}(\bm{x}_1)
(\eta_j(\bm{x}_2)-\eta_{j-1}(\bm{x}_2)),\\
\nonumber
p(z=0,y_1=j| \bm{x}_{1:2})=&\eta_j(\bm{x}_2)
(\eta_j(\bm{x}_1)-\eta_{j-1}(\bm{x}_1)),\\
\nonumber
p(z=0,y_2=j| \bm{x}_{1:2})=&\left(1- \eta_{j-1}(\bm{x}_1)\right)
(\eta_j(\bm{x}_2)-\eta_{j-1}(\bm{x}_2)).
\end{align}
\end{proposition}
\subsection{Ordinal triplet observation}
Ordinal triplet observation task is similar to triplet observation task, each bag has three instances (i.e. $m=3$) and the aggregate function is defined as $g(y_{1:3})=\mathbb{I}\left[d(y_1,y_2)<d(y_1,y_3)\right]$. In ordinal regression, we define $d(y_1,y_2)=|y_1-y_2|$.
\begin{proposition}
For the problem of classification via ordinal triplet comparison ($m=3$), $p(z| \bm{x}_{1:m})$ and $p(z,y_i=j| \bm{x}_{1:m})$ of $\mathcal{L}_{\mathrm{agg}}$ in Eq.~(\ref{L_agg}) can be empirically estimated by
\begin{align}
\nonumber
p(z=1| \bm{x}_{1:3})=& \sum_{j=1}^k\sum_{v=1}^k (\eta_v(\bm{x}_3)-\eta_{v-1}(\bm{x}_3)) p(|y_2-j|<|v-j|| \bm{x}_2)(\eta_j(\bm{x}_1)-\eta_{j-1}(\bm{x}_1))\\
\nonumber
p(z=v| \bm{x}_{1:3}) =& 1-p(z=1| \bm{x}_{1:3})
\end{align}

and

\begin{align}
\nonumber
p(z=1,y_1=j| \bm{x}_{1:3}) &= \sum_{v=1}^k (\eta_v(\bm{x}_3)-\eta_{v-1}(\bm{x}_3))
p(|y_2-j|<|v-j|| \bm{x}_2)(\eta_j(\bm{x}_1)-\eta_{j-1}(\bm{x}_1))\\
\nonumber
p(z=1,y_2=j| \bm{x}_{1:3}) &= \sum_{v=1}^k (\eta_v(\bm{x}_1)-\eta_{v-1}(\bm{x}_1))
p(|y_3-v|>|j-v||\bm{x}_3)(\eta_j(\bm{x}_2)-\eta_{j-1}(\bm{x}_2))\\
\nonumber
p(z=1,y_3=j|\bm{x}_{1:3}) &= \sum_{v=1}^k (\eta_v(\bm{x}_1)-\eta_{v-1}(\bm{x}_1))
p(|y_2-v|<|j-v||\bm{x}_2)(\eta_j(\bm{x}_3)-\eta_{j-1}(\bm{x}_3))\\
\nonumber
p(z=0,y_1=j|\bm{x}_{1:3}) &= \sum_{v=1}^k (\eta_v(\bm{x}_3)-\eta_{v-1}(\bm{x}_3))
(1-p(|y_2-j|<|v-j||\bm{x}_2))(\eta_j(\bm{x}_1)-\eta_{j-1}(\bm{x}_1))\\
\nonumber
p(z=0,y_2=j|\bm{x}_{1:3}) &= \sum_{v=1}^k (\eta_v(\bm{x}_1)-\eta_{v-1}(\bm{x}_1))
(1-p(|y_3-v|>|j-v||\bm{x}_3))(\eta_j(\bm{x}_2)-\eta_{j-1}(\bm{x}_2))\\
\nonumber
p(z=0,y_3=j|\bm{x}_{1:3}) &= \sum_{v=1}^k (\eta_v(\bm{x}_1)-\eta_{v-1}(\bm{x}_1))
(1-p(|y_2-v|<|j-v||\bm{x}_2))(\eta_j(\bm{x}_3)-\eta_{j-1}(\bm{x}_3))
\end{align}
For the calculation of $p(|y_2-v|<|j-v|| \bm{x}_2)$ and $p(|y_3-v|>|j-v|| \bm{x}_3)$ in above equation:

\begin{align}
\nonumber
p(|y_2-v|<|j-v|| \bm{x}_2)&=\mathrm{max}(\eta_{|j-v|+v-1}(\bm{x}_2)-\eta_{v-|j-v|}(\bm{x}_2),0)\\
\nonumber
p(|y_3-v|>|j-v|| \bm{x}_3)&=
1-\left(\eta_{v+|j-v|}(\bm{x}_3)-\eta_{v-|j-v|-1|}(\bm{x}_3)\right)
\nonumber
\end{align}
The function of $\mathrm{max}$ in the first equation is to make the equality holds when $|j-v|=0$.
\end{proposition}
\section{Experiments Details}
\label{Appendix_Experiments}
\subsection{Training Algorithm}

\begin{algorithm}[!t]
   \caption{RC Algorithm}
   \label{alg:rc}
   {\bfseries Input:} Model $f$, epoch $T_{\text{max}}$, iteration $I_{\text{max}}$, size of label space $k$, whether to use log-likelihood to initialize $\text{flag}_{\text{init}}$, log-likelihood initialize epoch $T_{\text{init}}$, whether to use confidence matrix $\text{flag}_{\text{mat}}$, aggregate observation training set $\mathcal{D}=\{(\bm{x}_{1:m}^{(i)},z^{(i)})\}_{i=1}^n$;
\begin{algorithmic}[1]
   \IF{$\text{flag}_{\text{mat}}$=TRUE}
   \STATE {\bfseries Initialize} confidence tensor $C\in \mathbb{R}^{n\times m \times k}$,$C_{ijv}=\frac{1}{k}, \forall 1\leq i \leq n,1\leq j\leq m,1\leq v \leq k$, we use $C_{ijv}$ to store $\eta_v(\bm{x}^{(i)}_j)$;
   \ENDIF
   \FOR{$t=1,2,\dots,T_{\text{max}}$}
   \STATE {\bfseries Shuffle} $\mathcal{D}=\{(\bm{x}_{1:m}^{(i)},z^{(i)})\}_{i=1}^n$;
   \FOR{$j=1,2,\dots I_{\text{max}}$ }
   \STATE {\bfseries Fetch} mini-batch $\mathcal{D}_j$ from $\mathcal{D}_j$;
   \IF{$\text{flag}_{\text{init}}$ and $t\leq T_{\text{init}}$}
   \STATE {\bfseries Update} model $f$ by log-likelihood method \cite{zhang2020learning};
   \ELSE
   \IF{$\text{flag}_{\text{mat}}$=True}
   \STATE {\bfseries Fetch} $\eta(\bm{x}_v^{(i)}), 1\leq v\leq m$ from $C$;
   \ELSE
   \STATE {\bfseries Calculate} $\eta(\bm{x}_v^{(i)}), 1\leq v\leq m$ by model $f$ and detach the gradient;
   \ENDIF
   \STATE {\bfseries Update} model $f$ by $\hat{R}(f)_{\mathrm{agg}}$ in Eq.~(\ref{empirical_risk});
   \ENDIF
   \IF{$\text{flag}_{\text{mat}}=$TRUE}
   \STATE {\bfseries Update} $C$ by model $f$;
   \ENDIF
   \ENDFOR
   \ENDFOR
\end{algorithmic}
   {\bf Output:} $f$.
\end{algorithm}
The pseudo-code of our proposed algorithm is presented in Algorithm \ref{alg:rc}. The training algorithm work with a similar process with RC \cite{feng2020provably} which treats $\eta(\bm{x})$ as weights and updates $\eta(\bm{x})$ during the training process. The main training process is provided between line 11 to line 19 in Algorithm \ref{alg:rc}. It contains 2 steps: 1) obtaining approximated $\eta(\bm{x})$. 2) using the approximated $\eta(\bm{x})$ to calculate $\hat{R}_{\mathrm{agg}}(f)$ and update model $f$. Since different CFAO tasks need different training strategies, we provide two strategies to obtain approximated $\eta(\bm{x})$. One way is to approximate $\eta(\bm{x})$ using the current model outputs. The other way is to approximate $\eta(\bm{x})$ by the model outputs from last epoch. We implement this by storing the outputs of model in a matrix and fetching $\eta(\bm{x})$ from the matrix during training. These two strategies corresponding to whether to use a radical way to update weights during training. We use $\text{flag}_{\text{mat}}$ to denote the strategy to update weights during training.  We use a matrix to store $\eta(\bm{x})$ when $\text{flag}_{\text{mat}}$ is set to TRUE, otherwise we obtain $\eta(\bm{x})$ by current model outputs. The matrix is initialized uniformly, which means we initialize all elements in the matrix to $\frac{1}{k}$.

$p(z | \bm{x}_{1:m})$ and $p(z,y_i=j | \bm{x}_{1:m})$ play an important rule in UUM to provide purified supervision for the classifier to learn. Especially, the value of these two probability functions depends on two components during the training process, i.e. $\eta(\bm{x})$ approximated by the model outputs and $z$ given by the aggregate observation dataset. During the warm-up phase, $p(z \mid \bm{x}_{1:m})$ and $p(z,y_i=j \mid \bm{x}_{1:m})$ mainly depend on $z$ since the model learned limited information in warm-up phase.

\begin{table*}[!b]
\caption{Test performance of UUM on $\text{flag}_{\text{init}}$=FALSE for pairwise similarity}
\label{UUM_similarity_noinit}
\resizebox{1.0\textwidth}{!}{
\setlength{\tabcolsep}{3.5mm}{
\begin{sc}
\begin{tabular}{c|ccccc}
\toprule
Dataset   & MNIST             & KUZUSHIJI           & FASHION             & CIFAR10           & SVHN               \\ \midrule
log-likelihood & $98.87\pm 0.06\%$  & $93.40\% \pm 0.64\%$      & $88.75 \pm 0.31\%$ & $71.45 \pm 0.47\% \bullet$ & $91.06 \pm 0.24\%\bullet$ \\
UUM(ours) & $\bm{99.01\pm 0.13\%}$ & $\bm{93.66 \pm 0.47\%} $ & $\bm{89.00 \pm 0.49 \%}$ & $\bm{73.45\pm 0.40\%}$ & $\bm{92.37\pm 0.23 \%}$\\ \bottomrule
\end{tabular}
\end{sc}
}
}
\end{table*}

However, in some CFAO problems, $z$ provides little information to weights during the warm-up phase, e.g. pairwise similarity and triplet comparison. If we set $\eta_j(\bm{x})=\frac{1}{k}$ for all $j$, $\frac{p(z,y_i=j\mid \bm{x}_{1:m})}{p(z\mid \bm{x}_{1:m})}$ would equal to $\frac{1}{k}$ for all $i$ and $j$ no matter $z$ takes 0 or 1, which means aggregate label would provides little information to help UUM approximating weights precisely. In order to obtain precise weights, we could use another method which dose not rely on weights to warm up model. We use log-likelihood as the warm-up method in our algorithm. The warm-up phase is shown in line 9 in Algorithm \ref{alg:rc}. We use $\text{flag}_{\text{init}}$ and $T_{\text{init}}$ to denote whether to use log-likelihood to warm up model and the number of warm-up epoch respectively.

We also conduct experiments on UUM on $\text{flag}_{\text{init}}=\text{FALSE}$ for pairwise similarity on benchmark datasets. We trained model for 100 epochs. For MINIST, Kuzushiji-MNIST and Fashion, the $\text{falg}_{\text{mat}}$ is set to FALSE in first 30 epoch and set to TRUE in last 70 epochs. For CIFAR-10 and SVHN, the $\text{flag}_{\text{mat}}$ is set to FALSE in first 50 epoch and set to TRUE in last 50 epochs. The experimetal results are provided on Table \ref{UUM_similarity_noinit}.


\subsection{Training Details}
We used Adam \cite{kingma2014adam} optimizer with 0 weight decay to train the model. The learning rates were 1e-3, 1e-3 and 2e-1 for benchmark datasets, UCI datasets and MIL datasets respectively. The batch size is 128 for benchmark datasets and UCI datasets. We search the batch size from (128, 256,512,1024,2048,4096) for MIL datasets. The model is trained for 100, 200, and 3500 epochs for benchmark datasets, UCI datasets and MIL datasets respectively. We evaluate test performance on the model obtained the best performance on validation sets.

The $\text{flag}_{\text{init}}$ in Algorithm \ref{alg:rc} is set TRUE for pairwise similarity and triplet comparison and set FALSE for MIL and LLP. $\text{flag}_{\text{mat}}$ is set TRUE for classification via pairwise similarity/triplet comparison/learning from label proportions, and set FALSE for multiple-instance learning. The value of $T_{\mathrm{init}}$ in pairwise similarity and triplet comparison is set to 20 and 100 for benchmark datasets and UCI datasets respectively.

\subsection{Benchmark Datasets for Multiple-Instance Learning}

We use five commonly used benchmark datasets in MIL studies \cite{dietterich1997solving,andrews2002support}, including {Musk1}, {Musk2}, {Elephant}, {Fox}, and {Tiger}. For these datasets, {Musk1} has 47 positive bags and 45 negative bags. {Musk2} consists of 39 positive bags and 63 negative bags. The other three datasets contain  100 positive bags and 100 negative bags. It is worth noting that these datasets are too small to evaluate the task of MIL from similar and dissimilar bags, we follow \citet{bao2018convex} to augment them for increasing the number of bags. Specifically, bags chosen randomly from the original datasets were duplicated
and then Gaussian noise with mean zero and variance 0.01 was
added to each dimension. In this way, we increased the number
of samples in the Musk datasets ({Musk1} and {Musk2}) 10 times and
the Corel datasets ({Elephant}, {Fox}, and {Tiger}) 5 times. Table \ref{benchmark_datasets} reports the characteristics of these datasets\footnote{ \url{http://www.cs.columbia.edu/~andrews/mil/datasets.html}} after preprocessing.
\begin{table*}[hb]
\centering
\caption{The characteristics of the used benchmark datasets for multiple-instance learning.}
\label{benchmark_datasets}
\resizebox{1.0\textwidth}{!}{
\setlength{\tabcolsep}{4.0mm}{
				\begin{tabular}{cccccc}
					\toprule
					Dataset & \# Features & \# Positive bags & \# Negative bags & \# Avg. Pos. Ins. per bag & \# Avg. Neg. Ins. per bag \\
					\midrule
					Musk1 & 166 & 475 & 445 & 2.2$\pm$2.5 & 2.9$\pm$7.0 \\
					Musk2 & 166 & 413 & 607 & 8.9$\pm$22.7 & 49.9$\pm$169.7 \\
					Elephat & 230 & 504 & 496 & 3.9$\pm$4.2 & 3.2$\pm$3.6 \\
					Fox & 230 & 498 & 502 & 3.2$\pm$3.6 & 3.4$\pm$3.8 \\
					Tiger & 230 & 506 & 494 & 2.8$\pm$3.1 & 3.4$\pm$3.9 \\
					\bottomrule
				\end{tabular}
				}
				}
\end{table*}

\section{Additional Experiments for Variant Size of Training Set}
Since the size of training sets may vary from a large range depending on different tasks in practice, we conducted additional experiments by reducing the sample size for training. The experimental results are shown in Table~\ref{variant_set_similarity} and Table~\ref{variant_set_tri}. As shown in Table~\ref{variant_set_similarity} and Table~\ref{variant_set_tri}, when reducing the sample size for training, our proposed method consistently outperforms other compared methods.

\begin{table*}[t]
\centering
\caption{Experiments on classification via pairwise similarity with different aggregate observation sample size formed by CIFAR10}
\label{variant_set_similarity}
\resizebox{1.0\textwidth}{!}{
\setlength{\tabcolsep}{4.0mm}{
\begin{tabular}{cccccc}
\toprule
Method               & 10000    & 30000    & 60000    & 90000    & 120000   \\
\midrule
log-likelihood & $33.30\% $&$ 56.15\% $&$ 66.30\% $&$ 70.97\% $&$ 71.45\%$ \\
siamese        &$ 31.74\% $&$ 41.63\% $&$ 48.01\% $&$ 54.27\% $&$ 58.16\%$ \\
contrastive    &$ 20.23\% $&$ 20.35\% $&$ 20.66\% $&$ 21.79\% $& $23.69\%$\\
\midrule
UUM            & $\bm{34.14\%}$ & $\bm{59.02\%}$ & $\bm{69.17\%}$ & $\bm{72.64\%}$ &$ \bm{72.52\%}$\\
\bottomrule
\end{tabular}
}
}
\end{table*}

\begin{table*}[t]
\centering
\caption{Experiments on classification via triplet comparison with different aggregate observation sample size formed by CIFAR10}
\label{variant_set_tri}
\resizebox{1.0\textwidth}{!}{
\setlength{\tabcolsep}{4.0mm}{
\begin{tabular}{cccccc}
\toprule
Method               & 10000    & 30000    & 60000    & 90000    & 120000   \\
\midrule
log-likelihood &$31.74\%$ & $53.15\%$ & $66.66\%$ & $68.67\%$ & $70.57\%$\\
triplet     &$26.48\%$ & $33.81\%$& $7.21\%$ & $39.85\%$ & $47.09\%$\\
(2+1)tuple  &$24.64\%$ & $29.74\%$& $36.65\%$ & $40.55\%$ & $45.53\%$\\
\midrule
UUM       &$\bm{33.03\%}$ & $\bm{56.44\%}$ & $\bm{63.38\%}$ & $\bm{72.97\%}$ & $\bm{72.09\%}$\\
\bottomrule
\end{tabular}
}
}
\end{table*}

\end{document}